\documentclass[preprint,nopreprintline,12pt]{elsarticle}
\usepackage[T1]{fontenc}
\usepackage[utf8]{inputenc}
\usepackage{lmodern}
\usepackage[a4paper,margin=20mm]{geometry}
\usepackage{amsmath,amssymb,booktabs,array,longtable}
\usepackage{graphicx}
\usepackage[section]{placeins}
\AddToHook{cmd/subsection/before}{\FloatBarrier}

\usepackage{caption}
\newcolumntype{L}[1]{>{\raggedright\arraybackslash}p{\dimexpr#1\linewidth-2\tabcolsep\relax}}
\newcolumntype{R}[1]{>{\raggedleft\arraybackslash}p{\dimexpr#1\linewidth-2\tabcolsep\relax}}
\biboptions{numbers,square,sort&compress}
\journal{}

\pdftrailerid{}
\PassOptionsToPackage{pdfcreator={},pdfproducer={}}{hyperref}

\usepackage[hidelinks]{hyperref}
\makeatletter
\AtBeginDocument{\let\Hy@author\@empty}
\makeatother

\begin{document}
\begin{frontmatter}

\title{\textbf{From MS/MS evidence to adaptive spectroscopic characterization: Reinforcement learning guided by candidate molecular structures}}

\author[affiliation,ucas]{\textbf{Bowen Gao}}
\ead{gaobowend@gmail.com}
\author[affiliation]{\textbf{Lei Zhu}}
\ead{leizhu@mail.sim.ac.cn}
\author[affiliation]{\textbf{Yiying Wang}\texorpdfstring{\corref{corresponding}}{}}
\ead{yywang@mail.sim.ac.cn}
\author[affiliation]{\textbf{Wenjie Yu}}
\ead{casan@mail.sim.ac.cn}
\address[affiliation]{State Key Laboratory of Materials for Integrated Circuits, Shanghai Institute of Microsystem and Information Technology, Chinese Academy of Sciences, Shanghai 200050, China}
\address[ucas]{University of Chinese Academy of Sciences, Beijing 100049, China}
\cortext[corresponding]{Corresponding author.}

\begin{abstract}
Small-molecule identification from tandem mass spectrometry (MS/MS) may require complementary measurements to distinguish competing structures, but acquiring every available spectrum increases experimental demands. Herein, we propose a structure-informed deep Q-network (SI-DQN) that uses existing MS/MS evidence to decide whether further characterization is needed, which measurement to request, and when to stop. The policy combines spectral representations and candidate rankings with score-weighted descriptors of the candidates' chemical features and structural differences. The policy was evaluated on 79,462 held-out samples with simulated MS/MS, infrared (IR), and nuclear magnetic resonance (NMR) spectra from the Multimodal Spectroscopic Dataset. Evaluation used fixed hard candidate sets containing the correct structure and equal normalized acquisition costs. Averaged across three training runs, SI-DQN increased Top-1 accuracy from 80.10\% to 93.78\% for positive-ion MS/MS starts and from 70.44\% to 93.31\% for negative-ion starts. The corresponding mean additional measurement counts were only 0.227 and 0.340 per sample, respectively. Complete acquisition achieved 97.72\% accuracy with four additions, whereas SI-DQN reduced the additional measurement count by 94.3\% and 91.5\%. Relative to a rule-based policy, SI-DQN reduced the composite loss combining terminal ranking error and acquisition cost by 19.3\% and 25.5\%. Adding structural information to DQN further improved accuracy while reducing requests across all three training runs. At comparable ranking uncertainty, SI-DQN increased request rates relative to DQN more strongly in groups with high structural disagreement, where larger reductions in composite loss were also observed. Together, these results support using the ambiguity among candidate structures after MS/MS analysis to guide further characterization, providing a quantitative basis for balancing identification performance against subsequent measurement effort.
\end{abstract}

\end{frontmatter}

\section{Introduction}
\label{sec:introduction}

Liquid chromatography--tandem mass spectrometry (LC--MS/MS) is widely adopted for small-molecule identification, as the fragment-ion patterns in MS/MS spectra provide rich information on molecular substructures and backbone connectivity~\cite{duhrkop2015searching,duhrkop2019sirius,wang2021cfmid4}.
However, practical structure identification often requires additional complementary spectroscopic measurements~\cite{alberts2024mssd,priessner2026mmst,gao2021graphyne,gao2021mlsynthesis,huang2025unipoly,gao2026robust}.
For instance, infrared (IR) and nuclear magnetic resonance (NMR) can help distinguish candidate structures through information about functional groups and local chemical environments~\cite{alberts2024infrared,hu2024accurate}. These additional measurements also require time and labor, increasing the demands on laboratory resources~\cite{beckonert2007metabolic,huang2021framework}. Therefore, once an MS/MS result is available, researchers must decide whether further characterization is needed and, if so, which measurement would provide the most useful additional information.

Reinforcement learning, combined with deep neural networks, offers a way to make these decisions using the evidence already available~\cite{mnih2015human,hwangbo2019learning}.
In particular, value-based methods (e.g., Q-learning) estimate how each choice affects the eventual identification result and the total acquisition cost~\cite{watkins1992qlearning,janisch2019classification}.
These estimates can therefore guide a sequence of spectroscopic measurements, with each new result informing whether to continue characterization or stop.
Recent studies have used reinforcement learning to optimize data acquisition within individual characterization techniques.
For example, Schloz et al. used deep reinforcement learning to optimize scan positions within electron ptychography, improving the structural similarity index by up to 15.8\% over sparse-grid scanning at the same measurement budget~\cite{schloz2023deep}.
Liu et al. combined ensemble deep kernel learning with a reinforcement learning policy to select predictive image channels and guide local spectroscopy within piezoresponse force microscopy. For polarization--voltage loop-area prediction in BiFeO$_3$, the amplitude-channel reward stabilized at 0.5--0.6 after approximately 50 exploration steps~\cite{liu2023learning}.
Tang et al. used a deep Q-network to select successive spectral bands within stimulated Raman scattering imaging. This approach achieved an average accuracy of 94.50\% for polymer segmentation at a 50\% band budget without characteristic-band priors, compared with 70.96\% for random sampling~\cite{tang2026overcoming}.
Reinforcement learning has also been used to infer molecular structures from \textsuperscript{13}C NMR spectra or combined IR and \textsuperscript{13}C NMR spectra through sequential molecular construction~\cite{sridharan2022deep,devata2024deepspinn}.
These studies focused on optimizing data acquisition within individual characterization techniques or inferring molecular structures from available spectra, but did not address how to select the next complementary characterization modality.
More importantly, for small-molecule identification beginning with MS/MS, a further challenge is to connect the chemical differences among competing candidates to the choice of complementary spectroscopic measurements and the decision to stop, while balancing ranking performance against acquisition cost.
Ranking scores summarize relative support for candidates but do not explicitly describe the functional-group or local chemical-environment differences that distinguish them.
Incorporating these differences into the decision state could help direct further measurements toward the remaining structural ambiguity.

Herein, we propose a structure-informed deep Q-network (SI-DQN), which combines value-based reinforcement learning with chemical domain knowledge to select spectroscopic measurements adaptively. Starting from MS/MS, the model dynamically decides whether to request additional IR, \textsuperscript{1}H NMR, or \textsuperscript{13}C NMR measurements, or to stop immediately and report the identification result. The model uses the available spectral evidence, the current candidate ranking, and the structural differences among competing candidates to make these decisions. When another measurement is selected, the resulting evidence and updated candidate ranking continue to guide the next decision. By relating the structural differences among candidate molecules to the potential value of another measurement, the approach uses existing spectra to guide further characterization while balancing identification performance against acquisition cost.

\section{Methods}
\label{sec:methods}

\subsection{Problem formulation}
\label{sec:problem-formulation}

\begin{figure}[!htbp]
\centering
\includegraphics[width=\linewidth,height=0.68\textheight,keepaspectratio]{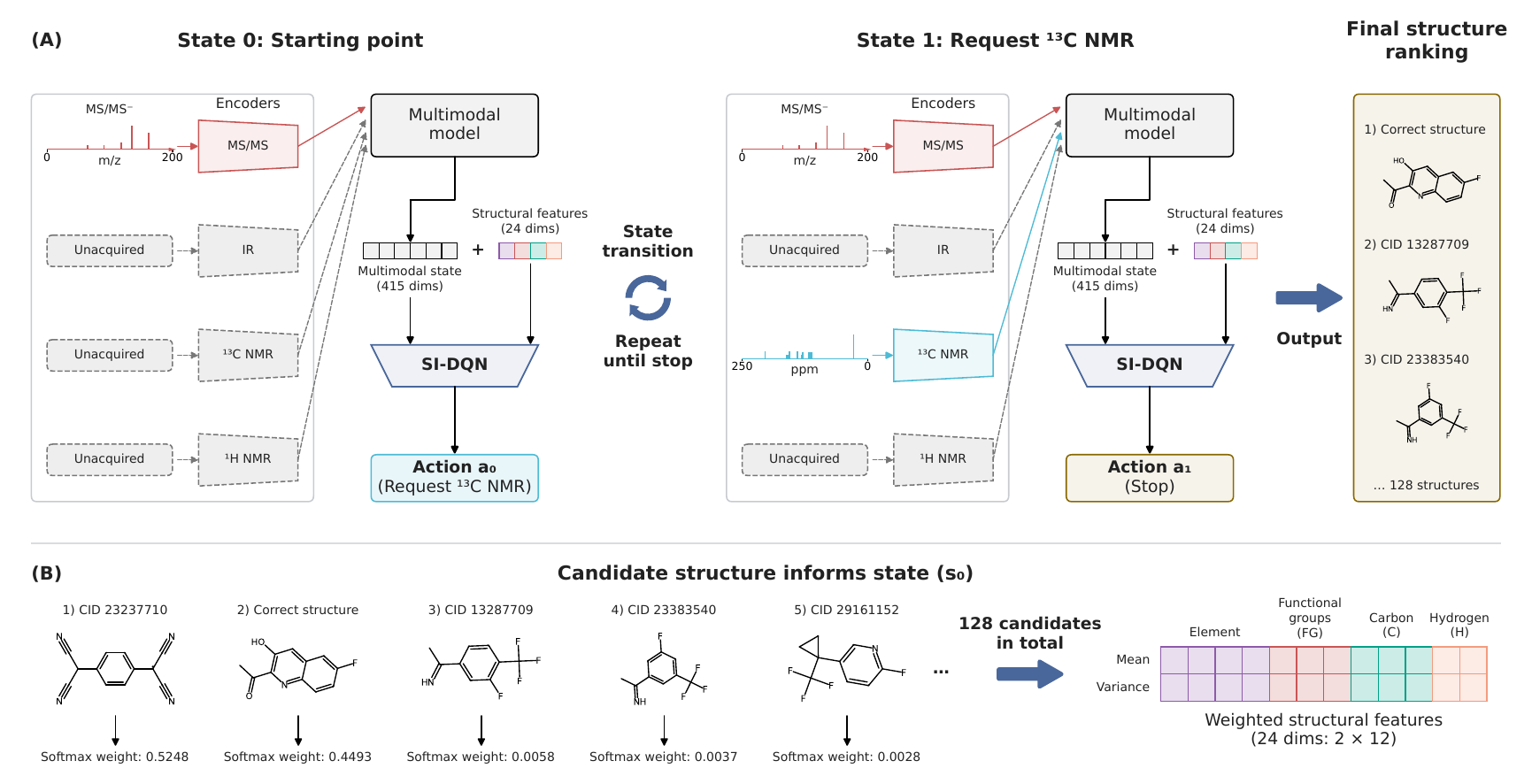}
\caption{Adaptive spectroscopic acquisition with SI-DQN. \textbf{(A)} A trajectory from MS/MS\textsuperscript{-} through \textsuperscript{13}C NMR acquisition, state updating, and stopping. Only observed spectra enter the frozen reranker. \textbf{(B)} Construction of score-weighted descriptor means and variances for the initial state, illustrated with the five highest-ranked candidates. Statistics use all 128 candidates: the correct structure and 127 structurally similar hard negatives. Correct-structure identities and ranks are post hoc annotations, excluded from policy inputs.}
\label{fig:1}
\end{figure}

Spectroscopic acquisition was formulated as a sequential decision problem, in which further measurements were selected on the basis of the available spectral evidence and candidate molecular structures. A structure-informed deep Q-network (SI-DQN), a deep reinforcement learning model, was developed to determine whether to request an additional measurement or stop acquisition. The dataset, data processing pipeline, and multimodal backbone were adopted from our previous multimodal molecular reranking study~\cite{gao2026robust}. Specifically, simulated spectra from the Multimodal Spectroscopic Dataset (MSSD)~\cite{alberts2024mssd} were processed using the same spectral preprocessing procedures. The dataset comprised 635,441 training samples, 79,440 validation samples, and 79,462 test samples. Each sample was assigned a fixed hard candidate set of up to 128 structures, comprising the correct structure and structurally similar alternatives. These candidates were reranked by the pretrained mixture-of-experts (MoE) backbone as additional spectral evidence became available. The backbone parameters and candidate sets remained fixed throughout policy training and evaluation.

Five channels covering four spectroscopic modalities were considered: IR, \textsuperscript{1}H NMR, \textsuperscript{13}C NMR, and positive- and negative-ion MS/MS (MS/MS\textsuperscript{+} and MS/MS\textsuperscript{-}) at 40 eV. At each decision step $t$, the state $s_t$ was constructed from the observed channels and the corresponding candidate scores. Channel availability, spectral representations, and candidate ranking information were included in the state, together with the structural features described in Section~\ref{sec:structural-features}. Each acquisition trajectory was initialized with one observed channel, and up to four additional measurements were permitted. Acquisition was terminated when Stop was selected, and the candidate ranking at that state was returned. The acquisition workflow and the construction of score-weighted structural summaries are illustrated in Fig.~\ref{fig:1}, where an actual case from the test set is followed through acquisition, state updating, and stopping.

For offline training and evaluation, 31 non-empty subsets of the five channels ($2^5-1$) were constructed for each sample. Each evidence subset represented a combination of observed channels. Backbone outputs and candidate ranks were precomputed for each subset, yielding a state graph in which each acquisition transition corresponds to the addition of one channel. A single policy was used across all five single-channel starting conditions, with particular emphasis on acquisition from an MS/MS starting point. SI-DQN was compared with Full, Rule, DQN, and Ideal to assess acquisition cost and identification performance, the contribution of structural information, and the identification--cost balance relative to an oracle reference.

Data eligibility and partitioning are described in Section~S1 and Table~S1 of the Supporting information. Spectral encoders, fusion, preprocessing, and candidate construction follow~\cite{gao2026robust}.

\subsection{Policy objective}
\label{sec:policy-objective}

A composite objective was defined to balance terminal ranking performance against the cost of additional spectroscopic measurements. The loss of an acquisition trajectory was defined as follows:
\begin{equation}
L_{\mathrm{task}}(\tau) = 1 - \frac{1}{r_T} + \frac{\lambda k}{4}
\label{eq:task-loss}
\end{equation}
where $r_T$ is the rank of the correct structure at the terminal step $T$, $k$ is the number of additional measurements in trajectory $\tau$, and $\lambda$ is the acquisition-cost weight. The denominator 4 is the number of channels available after the initial observation. With $\lambda=1$, each additional channel incurred an equal normalized cost of 0.25. The initial channel and Stop incurred no acquisition cost, and ranking loss was applied only at termination. The policy minimized expected trajectory loss, favoring acquisition when the expected reduction in ranking loss exceeded its cost.

Policy evaluation and model selection used mean trajectory loss, calculated for each evaluation population and starting condition as follows:
\begin{equation}
L = 1 - \mathrm{MRR} + 0.25\,\bar{n}
\label{eq:mean-trajectory-loss}
\end{equation}
Here, MRR is the mean reciprocal rank and $\bar{n}$ is the mean number of additional measurements. Lower values indicate a better balance between terminal ranking performance and acquisition cost. Top-1 accuracy was also reported. For one-step analyses, net gain was the increase in reciprocal rank after one request minus its normalized cost of 0.25. Network parameters were optimized using the temporal-difference loss described in Section~\ref{sec:double-dqn}.

Whole-cohort comparisons included all 79,462 test samples per MS/MS start. For learned policies, the three independent training runs were averaged equally within each sample before averaging over samples. One-step analyses instead weighted eligible sample--run records equally. Percentile 95\% confidence intervals were estimated by paired bootstrap resampling of dependency components (Section~S2.2)~\cite{efron1986bootstrap,field2007bootstrap}.

\subsection{Double DQN}
\label{sec:double-dqn}

\begin{table}[!htbp]
\centering
\small
\caption{Acquisition and stopping actions of DQN and SI-DQN.}
\label{tab:1}
\setlength{\tabcolsep}{4pt}
\renewcommand{\arraystretch}{1.16}
\begin{tabular}{@{}L{0.35}L{0.65}@{}}
\toprule
Action & Description \\
\midrule
Request IR & Add IR to observed evidence. \\
Request \textsuperscript{1}H NMR & Add \textsuperscript{1}H NMR to observed evidence. \\
Request \textsuperscript{13}C NMR & Add \textsuperscript{13}C NMR to observed evidence. \\
Request MS/MS\textsuperscript{+} & Add MS/MS\textsuperscript{+} (40 eV) to observed evidence. \\
Request MS/MS\textsuperscript{-} & Add MS/MS\textsuperscript{-} (40 eV) to observed evidence. \\
Stop & Stop acquisition and return the current ranking. \\
\bottomrule
\end{tabular}
\end{table}

Double DQN was adopted to estimate action values for sequential acquisition and stopping~\cite{mnih2015human,vanhasselt2016deep}. Here, $Q^{\pi}(s,a)$ denotes the expected remaining composite loss after action $a$ is taken in state $s$ and policy $\pi$ is followed thereafter. It includes subsequent acquisition costs and terminal ranking loss, so actions were selected by minimizing the estimated value. The deep neural network estimated one value for each action in Table~\ref{tab:1}.

Requests for channels that had already been observed were masked and could not be selected. The admissible action set $A(s)$ therefore contained four requests and Stop at each single-channel starting state, and only Stop after all five channels had been acquired.

The optimal action values satisfy separate Bellman equations for requesting a measurement and stopping. For an admissible request, the value combines the acquisition cost with the minimum expected loss from the next state:
\begin{equation}
Q^{*}(s,a_{\mathrm{request}}) = \frac{\lambda}{4} + \gamma\,\mathbb{E}\!\left[\min_{b\in A(s^{\prime})} Q^{*}(s^{\prime},b) \,\middle|\, s,a_{\mathrm{request}}\right]
\label{eq:bellman-request}
\end{equation}
where $Q^{*}$ is the minimum expected remaining composite loss, $a_{\mathrm{request}}$ acquires one unobserved channel, and $s^{\prime}$ is the resulting evidence state. The expectation is over successor states, with $b$ indexing their admissible actions. The immediate cost is $\lambda/4$, and $\gamma=1$ gives an undiscounted finite-horizon objective.

For Stop, no further acquisition cost is incurred, and the value is the expected ranking loss at the current state:
\begin{equation}
Q^{*}(s,\mathrm{Stop}) = \mathbb{E}\!\left[1-\frac{1}{r_s}\,\middle|\,s\right]
\label{eq:bellman-stop}
\end{equation}
where $r_s$ is the rank of the correct structure in state $s$.

An online network $Q_{\theta}$ and a target network $Q_{\bar{\theta}}$ with identical architectures were used. Following Double DQN, the online network selected the admissible action with the lowest predicted loss at the next state. The target network then estimated the remaining loss for that selected action~\cite{vanhasselt2016deep}. For a request, the training target was the acquisition cost plus this successor loss estimate. For Stop, the training target was the ranking loss at the current state. Ranks of the correct structure were used to calculate training targets and evaluation metrics, and were excluded from policy inputs.

The online network minimized the mean squared temporal-difference error between its predicted action values and the training targets. Squared errors were averaged first over admissible actions within each sampled state and then over states. Targets were held constant during backpropagation. The target network was updated by periodically copying the online parameters, and only the online network was used at inference. Training settings and deterministic selection among tied actions are specified in Section~S5.

\subsection{Structural features}
\label{sec:structural-features}

To support DQN in selecting suitable actions based on the available spectral evidence, chemical domain knowledge was incorporated through molecular descriptors of the candidate structures. DQN used a 415-dimensional multimodal state representation combining information about the observed channels and the corresponding candidate scores (Table~S2). The state included channel availability, candidate weights ordered by rank, and summary statistics of these weights. It also included the 256-dimensional fused spectral embedding, reliability and conflict indicators, and mixture-of-experts routing and fusion information. Candidate weights were derived from the backbone scores. This state was augmented with score-weighted statistics of the molecular descriptors to form the SI-DQN input.

For each valid candidate $j$, 12 molecular descriptors were calculated using RDKit and predefined substructure patterns. The four descriptor categories are summarized in Table~\ref{tab:2}.

\begin{table}[!htbp]
\centering
\small
\caption{Candidate molecular descriptors used for structural summaries.}
\label{tab:2}
\setlength{\tabcolsep}{4pt}
\renewcommand{\arraystretch}{1.16}
\begin{tabular}{@{}L{0.28}R{0.18}L{0.54}@{}}
\toprule
Chemical category & Number of\par descriptors & Descriptors \\
\midrule
Elemental composition & 4 & N; O; S; halogens \\
Functional groups & 3 & Carbonyl; hydroxyl; nitrile \\
Carbon environments & 3 & Aromatic, aliphatic sp\textsuperscript{3},\par and heteroatom-bonded carbons \\
Hydrogen environments & 2 & Hydrogens on (non)aromatic carbons \\
\bottomrule
\end{tabular}
\par\medskip
\begin{minipage}{\linewidth}\footnotesize
Descriptors are atom counts. Hydrogen descriptors count only carbon-bound hydrogens. Detailed counting rules are given in Table~S3.
\end{minipage}
\end{table}

Candidate scores were converted into weights by applying Softmax over the valid candidates. The score-weighted mean of each descriptor was defined as follows:
\begin{equation}
\mu_d(s) = \sum_{j\in C} w_j(s)f_{jd}
\label{eq:descriptor-mean}
\end{equation}
where $\mu_d(s)$ is the weighted mean of descriptor $d$ at evidence state $s$. The set $C$ contains the valid candidate molecular structures, indexed by $j$. The weight $w_j(s)$ is the Softmax weight derived from candidate $j$'s backbone score, and $f_{jd}$ is its value of descriptor $d$. The mean summarizes the descriptor values favored by the current candidate scores.

The corresponding centered weighted variance was defined as follows:
\begin{equation}
v_d(s) = \sum_{j\in C} w_j(s)\left[f_{jd}-\mu_d(s)\right]^2
\label{eq:descriptor-variance}
\end{equation}
where $v_d(s)$ measures disagreement in descriptor $d$ among candidate molecular structures under the current score weighting.

The 12 means followed by the 12 variances formed a 24-dimensional vector, which was standardized using statistics estimated only from the policy-training data. Appending the resulting structural features to the 415-dimensional multimodal state yielded the 439-dimensional structure-informed state used as input to SI-DQN.

After each acquisition, candidate scores and descriptor statistics were updated. Molecular descriptors remained fixed, while their contributions to the state changed with the available spectra.

The complete specification of the multimodal state, descriptor counting rules, and feature standardization is provided in Section~S3.

\subsection{Reference methods}
\label{sec:reference-methods}

SI-DQN was benchmarked against four reference methods using identical backbone outputs, candidate sets, evidence states, and a common composite-loss objective:

\begin{enumerate}
\renewcommand{\labelenumi}{\arabic{enumi})}
\item The full acquisition policy (denoted as Full) acquired all remaining channels to provide a reference for identification performance and acquisition cost under complete spectral evidence;
\item The rule-based acquisition policy (denoted as Rule) used a fixed acquisition order and a candidate-weight threshold for stopping, both selected on the designated validation subset;
\item The deep Q-network policy (denoted as DQN) was trained using the same Double DQN procedure with the 415-dimensional multimodal state to assess the contribution of the 24 structural features in SI-DQN;
\item The ideal acquisition policy (denoted as Ideal) used exact dynamic programming with ranks of the correct structure for all evidence subsets, including unacquired subsets, to establish an oracle lower bound on composite loss.
\end{enumerate}

Among these reference methods, Full, Rule, and Ideal required no Q-network training. Rule selection and the exact optimization procedure for Ideal are described in Section~S4.

\subsection{Training protocol}
\label{sec:training-protocol}

The data partitions were retained from the previous multimodal molecular reranking study~\cite{gao2026robust}. Training data were used for policy optimization and feature normalization. Separate validation subsets were used for neural checkpoint selection and Rule configuration, and the test set was reserved for final evaluation.

SI-DQN and DQN were each trained from random initialization in three independent runs to assess variability due to stochastic training. Each Double DQN update used 512 independently sampled state groups from the offline graph. Each group comprised one state and all its admissible actions, with a training target for each action. Each trained model was applied to all five starting conditions.

Checkpoint selection and early stopping used validation composite loss, averaged over the five starts within each sample and then over samples. The best saved checkpoint from each run was used for test evaluation. Sampling, optimization, and early-stopping settings are given in Section~S5 and Table~S4.

For the stratified analyses, thresholds for ranking uncertainty and structural disagreement were fixed before test analysis (Section~S6.3; Table~S5).

\section{Results and Discussion}
\label{sec:results}

\subsection{Identification performance and acquisition cost}
\label{sec:performance-cost}

\begin{figure}[!htbp]
\centering
\includegraphics[width=\linewidth,height=0.68\textheight,keepaspectratio]{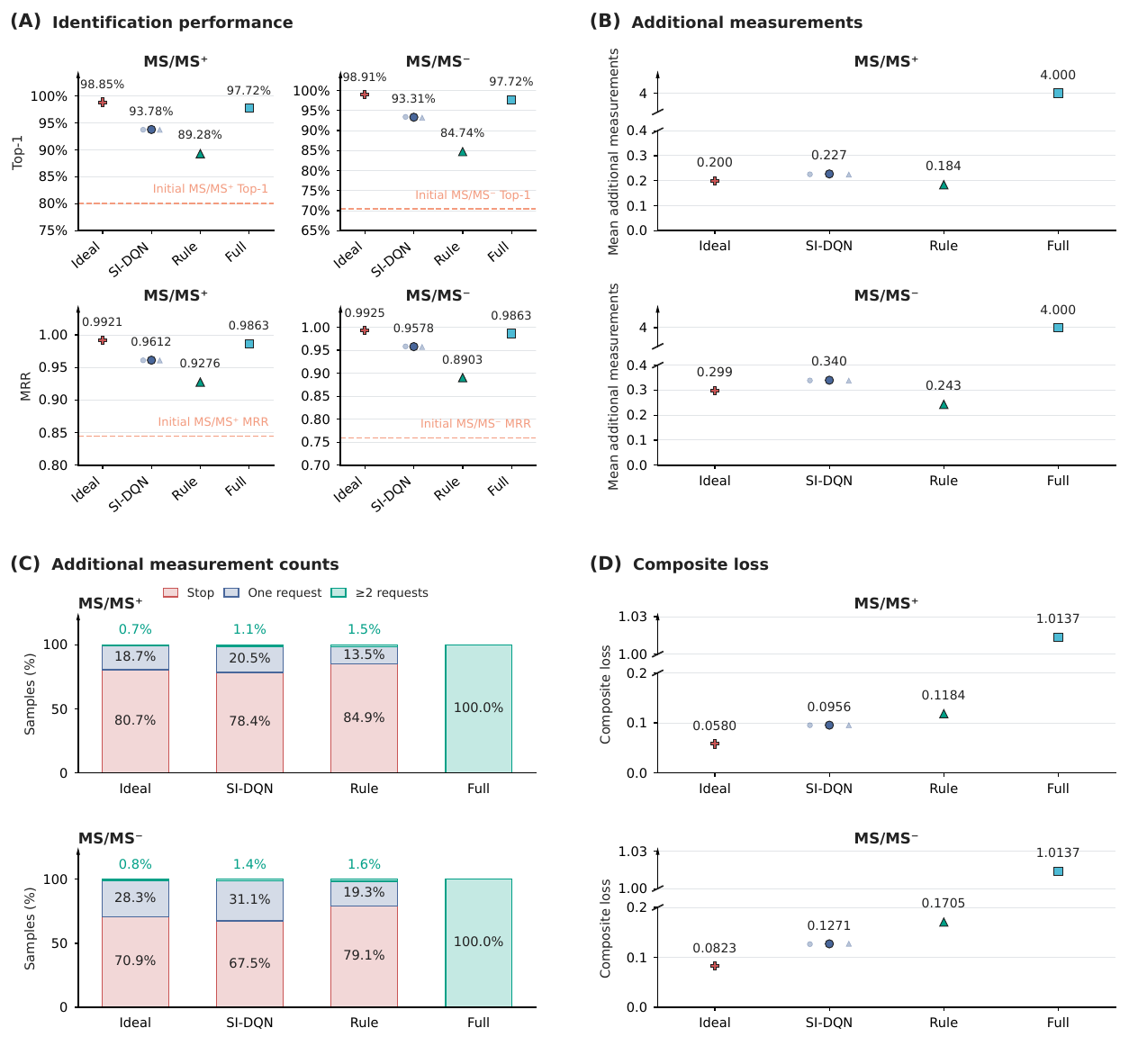}
\caption{Identification performance and acquisition cost of SI-DQN and reference policies with hard candidates from both MS/MS starts. \textbf{(A)} Terminal Top-1 accuracy and mean reciprocal rank (MRR). \textbf{(B)} Mean additional measurements. \textbf{(C)} Acquisition-depth distributions. \textbf{(D)} Composite loss. Large symbols show cohort means; pale symbols show the three SI-DQN training runs.}
\label{fig:2}
\end{figure}

Overall, selective acquisition substantially improved molecular identification from both MS/MS starting conditions while requiring only a small number of additional measurements. With SI-DQN, Top-1 accuracy increased relative to the initial MS/MS-only baseline, from 80.10\% to 93.78\% for MS/MS\textsuperscript{+} and from 70.44\% to 93.31\% for MS/MS\textsuperscript{-} (Fig.~\ref{fig:2}A), corresponding to gains of 13.68 and 22.87 percentage points, respectively. MRR increased in parallel, from 0.8444 to 0.9612 for MS/MS\textsuperscript{+} and from 0.7593 to 0.9578 for MS/MS\textsuperscript{-}, equivalent to relative increases of 13.8\% and 26.1\%. The larger improvement from MS/MS\textsuperscript{-} brought its terminal identification performance close to that obtained from MS/MS\textsuperscript{+}, showing the value of complementary acquisition across both starting conditions.

Notably, these improvements required a mean of only 0.227 and 0.340 additional measurements per sample for MS/MS\textsuperscript{+} and MS/MS\textsuperscript{-}, respectively (Fig.~\ref{fig:2}B). Compared with the four additions required by Full, SI-DQN reduced the number of additional measurements by 94.3\% and 91.5\%. Full achieved a higher Top-1 accuracy of 97.72\% and an MRR of 0.9863 with complete spectral evidence. SI-DQN therefore retained high identification performance with substantially fewer measurements. Most trajectories stopped immediately or after a single additional measurement (Fig.~\ref{fig:2}C), showing that this balance was achieved through selective, short acquisition trajectories. Overall metrics are provided in Table~S6.

The comparison with Rule further demonstrated the value of allocating additional measurements adaptively. Rule requested fewer additional measurements, averaging 0.184 and 0.243 per sample, with Top-1 accuracies of 89.28\% and 84.74\% for MS/MS\textsuperscript{+} and MS/MS\textsuperscript{-}, respectively. With slightly more measurements, SI-DQN improved Top-1 accuracy by 4.50 and 8.58 percentage points. When terminal ranking and acquisition cost were considered jointly, composite loss decreased from 0.1184 to 0.0956 for MS/MS\textsuperscript{+} and from 0.1705 to 0.1271 for MS/MS\textsuperscript{-} (Fig.~\ref{fig:2}D), corresponding to relative reductions of 19.3\% and 25.5\%. Thus, the ranking improvements achieved by the additional requests outweighed their acquisition cost under the common objective.

Finally, comparison with Ideal quantified the improvement relative to the oracle lower bound on composite loss. SI-DQN closed 37.8\% of the Rule-to-Ideal loss gap for MS/MS\textsuperscript{+} and 49.2\% for MS/MS\textsuperscript{-} (Fig.~\ref{fig:2}D; Table~S7). Together, these results establish that SI-DQN improved the identification--cost balance over Rule, with the larger benefit observed from the MS/MS\textsuperscript{-} starting condition. Notably, Ideal achieved slightly higher terminal accuracy than Full, as it used the known correct structure to guide acquisition and stopping. The following analysis examines how candidate structural information contributed to this allocation of measurement effort.

\pagebreak[0] 
\subsection{Contribution of structural information}
\label{sec:structural-contribution}

\begin{figure}[!htbp]
\centering
\includegraphics[width=\linewidth,height=0.68\textheight,keepaspectratio]{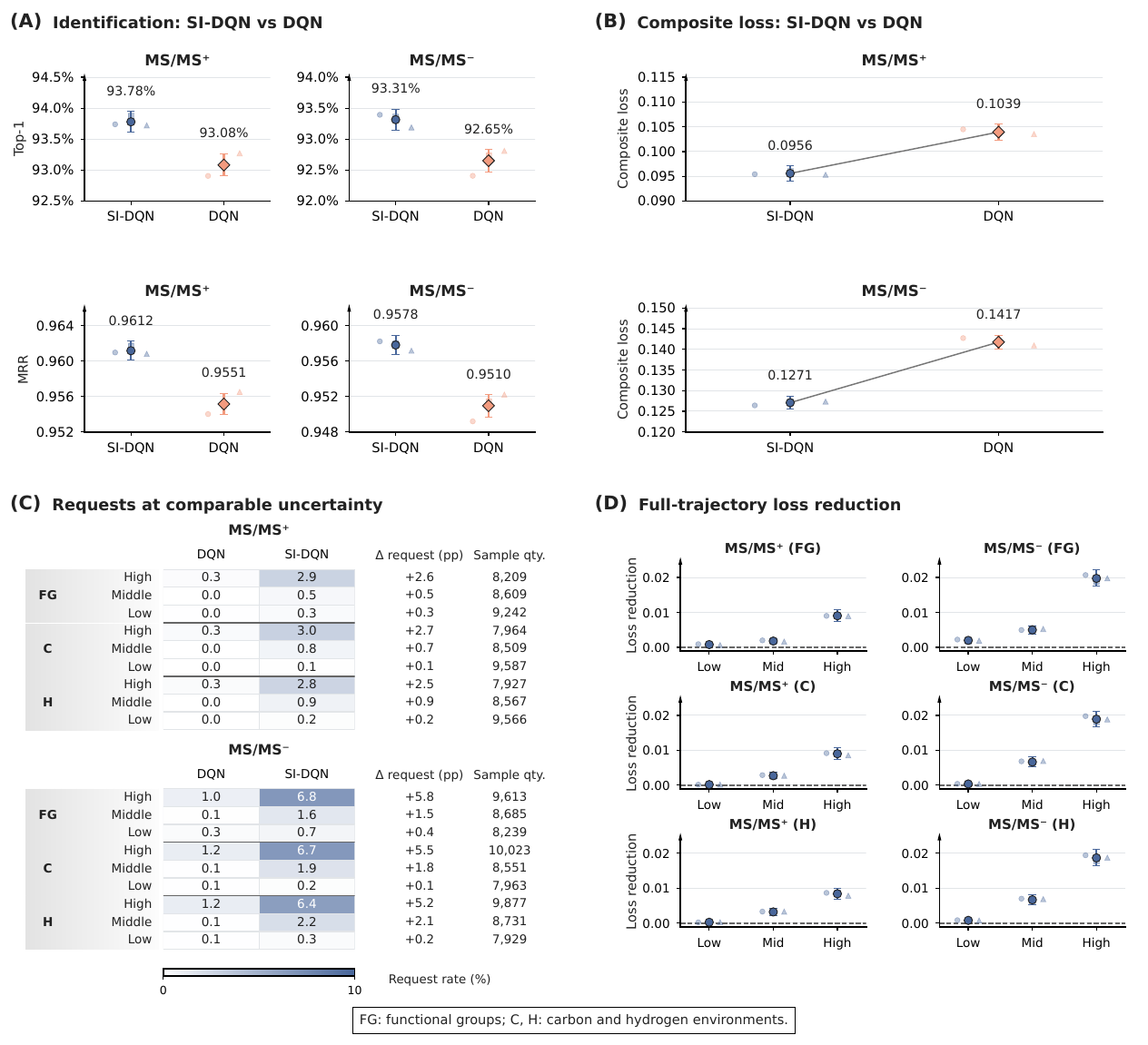}
\caption{Contribution of candidate structural information to acquisition decisions. \textbf{(A)} Terminal identification performance and \textbf{(B)} composite loss for DQN and SI-DQN. \textbf{(C)} Initial request rates stratified by structural disagreement within the middle tertile of normalized candidate-weight entropy. Adjacent columns report SI-DQN minus DQN differences in percentage points and sample counts. \textbf{(D)} Paired reduction in composite loss, $L_{\mathrm{DQN}}-L_{\mathrm{SI-DQN}}$, in the same groups. Positive values favor SI-DQN. Group definitions are given in Section~S6.3. Large points show means over samples; pale points show individual training runs. Error bars show paired-bootstrap 95\% confidence intervals.}
\label{fig:3}
\end{figure}

To determine whether candidate structural information contributed directly to acquisition decisions, SI-DQN was compared with DQN under the same reinforcement-learning framework. SI-DQN increased Top-1 accuracy by 0.70 percentage points for MS/MS\textsuperscript{+} and 0.66 percentage points for MS/MS\textsuperscript{-}, accompanied by MRR gains of 0.0060 and 0.0069, respectively (Fig.~\ref{fig:3}A). The direction of improvement was consistent across all three training runs. Specifically, the run-level Top-1 gains ranged from 0.45 to 0.83 percentage points for MS/MS\textsuperscript{+} and from 0.38 to 0.99 percentage points for MS/MS\textsuperscript{-}. These consistent gains indicate that the structural features provided information beyond that already contained in the multimodal state.

These identification gains were accompanied by simultaneous reductions in composite loss and acquisition cost. The paired loss reductions were 0.0083 for MS/MS\textsuperscript{+} and 0.0147 for MS/MS\textsuperscript{-} (Fig.~\ref{fig:3}B; Table~S7). Mean additional measurement counts decreased from 0.236 to 0.227 and from 0.371 to 0.340, respectively (Fig.~S1; Section~S6.1), corresponding to reductions of 3.9\% and 8.4\%. The decrease was observed in every training run. Therefore, the added structural information improved not only terminal ranking but also the efficiency with which measurement effort was allocated.

To further investigate why these structural summaries could support acquisition, score-weighted descriptor means were evaluated under actual and randomized score--structure pairings. Specifically, the candidates and their scores were kept unchanged while the score--descriptor pairings were randomly shuffled, in order to investigate whether the correct structure's descriptor values were better approximated by the weighted means under actual pairing. Across all four chemical categories, the means under actual pairing were closer to the corresponding descriptor values of the correct structure than those under randomized pairing (Fig.~S2A; Section~S6.2). For example, the mean normalized functional-group error was 0.134 under the actual pairing versus 0.747 under random pairing for MS/MS\textsuperscript{+}, and 0.217 versus 0.741 for MS/MS\textsuperscript{-}. In other words, random pairing produced functional-group errors approximately 5.6 and 3.4 times larger, respectively. Actual pairing also retained an advantage across all categories among samples with initially incorrect Top-1 predictions (Fig.~S2B). Within this more difficult subset, functional-group errors were 0.510 versus 0.605 for MS/MS\textsuperscript{+} and 0.586 versus 0.634 for MS/MS\textsuperscript{-}. Thus, the candidate distribution retained coarse chemical constraints even when the correct structure was not ranked first, providing chemically informative summaries at states where further evidence could improve identification.

Next, the behavioral analysis examined whether SI-DQN used this information to distinguish acquisition needs among samples with comparable ranking uncertainty. For this analysis, ranking uncertainty was measured using the normalized Shannon entropy of candidate weights~\cite{shannon1948mathematical}. The middle entropy tertile covered the middle third of the reference distribution, using thresholds fixed before test analysis (Section~S6.3). Within this tertile, SI-DQN requested additional evidence more frequently than DQN when structural disagreement was high (Fig.~\ref{fig:3}C). For high functional-group disagreement, the request rate increased from 0.3\% to 2.9\% for MS/MS\textsuperscript{+} and from 1.0\% to 6.8\% for MS/MS\textsuperscript{-}, representing 9.7- and 6.8-fold increases. By contrast, the corresponding increases in the low-disagreement groups were only 0.3 and 0.4 percentage points. Carbon- and hydrogen-environment disagreement showed similar patterns. These results indicate that structural summaries helped distinguish acquisition needs among samples with similar ranking uncertainty but different levels of structural disagreement. Candidate structural information therefore supplied an additional basis for assessing the potential value of another measurement.

The outcome analysis further showed that these localized changes in acquisition behavior were associated with larger improvements where structural disagreement was high. In the high functional-group disagreement group, the paired loss reduction was 0.0090 for MS/MS\textsuperscript{+} and 0.0197 for MS/MS\textsuperscript{-}, compared with only 0.0008 and 0.0020 in the corresponding low-disagreement groups (Fig.~\ref{fig:3}D). The high-disagreement reductions were therefore approximately 11.3 and 9.9 times larger. Carbon- and hydrogen-environment analyses showed the same ordering of point estimates, with the clearest gains again observed in the high-disagreement groups. Taken together with the overall measurement savings in Fig.~S1, these results indicate two complementary effects of structural information: (1) fewer additional measurements were requested across the cohort while terminal identification improved; and (2) request rates increased more strongly in groups with high structural disagreement, where the resulting trajectories showed larger loss reductions. This combination supports the use of candidate structural information to guide the allocation of measurement effort. Structural summaries connect uncertainty in the ranking to the chemical distinctions that complementary spectra may help resolve, providing a basis for selective acquisition beyond the candidate scores alone.

\subsection{Acquisition and stopping decisions}
\label{sec:acquisition-stopping}

Having established that structural information improved the allocation of measurement effort, the next question was how this allocation translated into acquisition depth and stopping behavior. SI-DQN stopped immediately for 78.4\% of MS/MS\textsuperscript{+} trajectories and 67.5\% of MS/MS\textsuperscript{-} trajectories, whereas Rule stopped for 84.9\% and 79.1\%, respectively (Fig.~\ref{fig:2}C). Thus, SI-DQN was more willing to initiate acquisition, particularly from the less accurate MS/MS\textsuperscript{-} starting condition. This behavior was consistent with the larger identification improvement observed from MS/MS\textsuperscript{-} in Fig.~\ref{fig:2}A. Most trajectories ended at the initial MS/MS state, concentrating further characterization on a minority of samples.

When acquisition was initiated, one additional measurement was usually sufficient for SI-DQN to terminate. One-addition trajectories accounted for 20.5\% of all MS/MS\textsuperscript{+} samples and 31.1\% of all MS/MS\textsuperscript{-} samples, compared with 13.5\% and 19.3\% under Rule. These trajectories were therefore approximately 1.52 and 1.61 times as common under SI-DQN. In contrast, the proportions requiring at least two additions were slightly lower under SI-DQN, at 1.1\% versus 1.5\% for MS/MS\textsuperscript{+} and 1.4\% versus 1.6\% for MS/MS\textsuperscript{-} (Fig.~\ref{fig:2}C). Accordingly, the greater acquisition effort relative to Rule was concentrated primarily in giving a single complementary measurement to more samples. This short-trajectory pattern explains how SI-DQN combined identification gains with low measurement requirements: after complementary evidence was incorporated, the updated candidate ranking supported a new decision to continue or stop. Reassessment after each request allowed acquisition depth to adapt to the evidence available for each sample.

Comparison with Ideal further highlighted the importance of coordinating acquisition and stopping. Ideal stopped immediately for 80.7\% of MS/MS\textsuperscript{+} trajectories and 70.9\% of MS/MS\textsuperscript{-} trajectories while achieving higher terminal identification performance (Fig.~\ref{fig:2}A--C). This reference illustrates the value of identifying samples that benefit from additional evidence and selecting an effective complementary measurement for those samples. SI-DQN coordinated these decisions by estimating the remaining composite loss for each request and Stop from the current evidence state. Its predominantly short trajectories show how this formulation supported selective acquisition from MS/MS starting states. Stopping was an integral part of acquisition planning, allowing resources to be concentrated on samples for which further characterization was expected to improve the identification--cost balance.

\subsection{Selection of complementary measurements}
\label{sec:complementary-measurements}

\begin{figure}[!htbp]
\centering
\includegraphics[width=\linewidth,height=0.68\textheight,keepaspectratio]{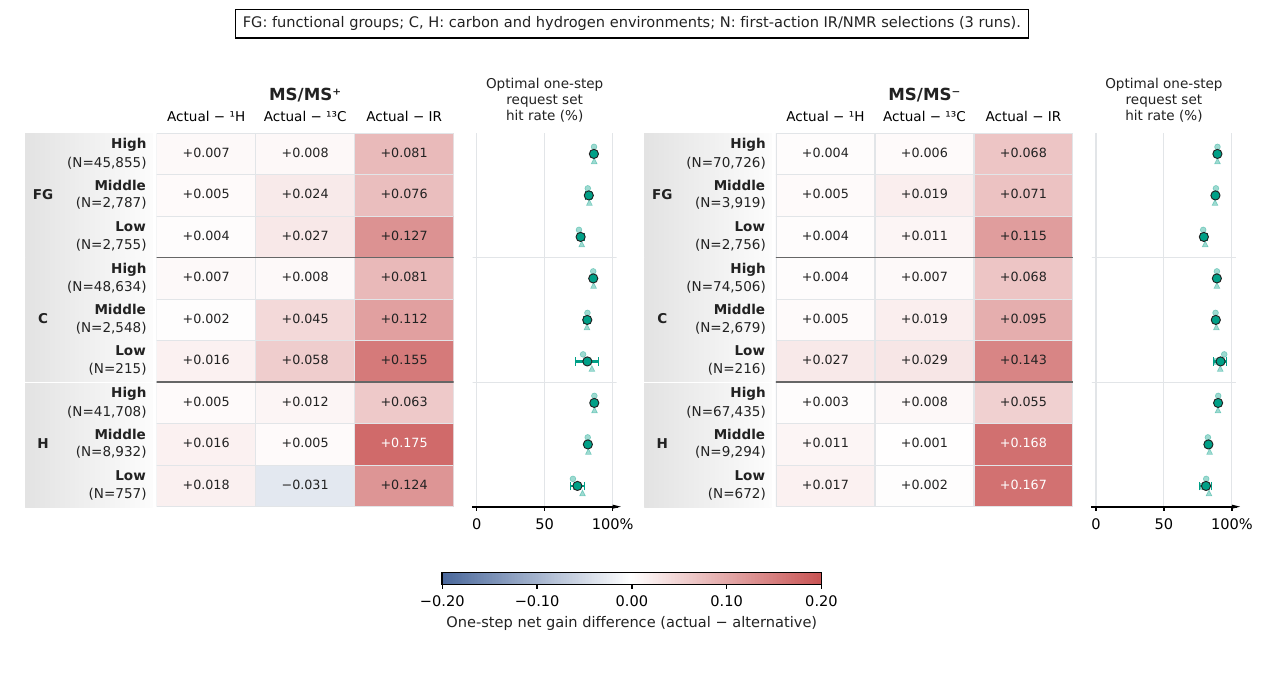}
\caption{One-step evaluation of SI-DQN measurement selection by structural disagreement. Heatmaps show mean differences in net gain between the actual request and each IR or NMR alternative. Net gain is the increase in the correct structure's reciprocal rank minus 0.25. Adjacent plots show the fraction of selections attaining the highest net gain among all four unobserved channels, including the opposite MS/MS polarity. All tied requests belong to this optimal one-step request set (Section~S6.3). $N$ counts eligible sample--run records (one sample evaluated by one trained model) with initial IR or NMR requests across all entropy levels. Large points show pooled estimates, pale points individual runs, and error bars paired-bootstrap 95\% confidence intervals.}
\label{fig:4}
\end{figure}

The acquisition-depth analysis established when SI-DQN tended to request additional evidence. The next question was whether the selected measurement was an appropriate complement for the particular sample. Acquisition was mainly concentrated on NMR. Among first requests from MS/MS\textsuperscript{+}, \textsuperscript{1}H NMR and \textsuperscript{13}C NMR accounted for 50.4\% and 49.3\%, respectively. From MS/MS\textsuperscript{-}, the balance shifted toward \textsuperscript{13}C NMR, with corresponding proportions of 44.0\% and 54.9\%. On the other hand, IR was requested less frequently. These frequencies identify the measurements favored by the learned policy. The one-step outcome comparisons in Fig.~\ref{fig:4} further assess whether those choices improved discrimination among the current candidates.

First, the clear separation in Fig.~\ref{fig:4} was observed between the actual selections and IR alternatives. Actual selections produced higher mean one-step net gains in every displayed structural-disagreement group, with differences ranging from approximately 0.055 to 0.175. For example, in the high functional-group disagreement groups, the advantages were 0.081 for MS/MS\textsuperscript{+} and 0.068 for MS/MS\textsuperscript{-}. Positive differences were also observed across the carbon- and hydrogen-environment groups. The strong preference for NMR is consistent with its ability to characterize local chemical environments that may remain ambiguous after MS/MS-based ranking~\cite{priessner2026mmst,wang2017identification}. The advantage also extended to functional-group disagreement strata. These observations support evaluating complementarity according to how effectively an added measurement separates the current candidate structures. A structural-disagreement category summarizes distinctions among candidates, while the usefulness of a spectrum depends on its contribution to resolving those distinctions in the context of the available evidence.

Second, the comparison among NMR measurements showed how SI-DQN refined this broad preference at the sample level. In the high functional-group disagreement groups, the mean one-step net gains of actual selections were approximately 0.007 and 0.004 higher than those of \textsuperscript{1}H NMR for MS/MS\textsuperscript{+} and MS/MS\textsuperscript{-}, respectively. The corresponding advantages over \textsuperscript{13}C NMR were approximately 0.008 and 0.006 (Fig.~\ref{fig:4}). These gains indicate that selecting an informative NMR measurement delivered much of the benefit, with sample-specific choice providing an additional improvement. However, the comparisons varied across structural-disagreement groups. In the low hydrogen-environment disagreement group from MS/MS\textsuperscript{+}, confidence intervals spanned zero for both NMR alternatives (Table~S7). These results show how candidate-specific measurement selection can build on the complementary information provided by NMR.

Third, selected measurements were compared with the optimal one-step request set, defined by the highest net gain among all four unobserved channels, including the opposite MS/MS polarity. Multiple requests could belong to this set if they attained the same best rank for the correct structure. Across the high-disagreement groups, approximately 86\% of actual selections from MS/MS\textsuperscript{+} and 89--90\% from MS/MS\textsuperscript{-} belonged to this set (Fig.~\ref{fig:4}). The high fraction of selections in the optimal set complements the mean-gain comparisons by showing how frequently an effective measurement was chosen for the individual sample. This distinction is useful for acquisition planning, where several spectroscopic measurements may provide alternative routes to the same ranking outcome.

Finally, channel requests were randomly reassigned among requesting samples to assess the benefit of matching measurements to individual samples. The requesting samples and request counts for each channel were held fixed. Actual one-step net gains were 0.2769 versus 0.2682 under random pairing for MS/MS\textsuperscript{+} and 0.3478 versus 0.3410 for MS/MS\textsuperscript{-} (Fig.~S3A--B; Section~S6.4). The paired advantages were 0.0086 and 0.0068, with positive differences in all three training runs (Table~S7). These comparisons identified a strong overall preference for measurements, particularly NMR, that complemented the MS/MS starting state, together with a further consistent gain from assigning those measurements to specific samples. Alongside the acquisition and stopping analysis, the results support planning complementary characterization through both the choice of samples to investigate and the choice of evidence to obtain.

\subsection{Molecular case studies}
\label{sec:molecular-cases}

\begin{figure}[!htbp]
\centering
\includegraphics[width=\linewidth,height=0.68\textheight,keepaspectratio]{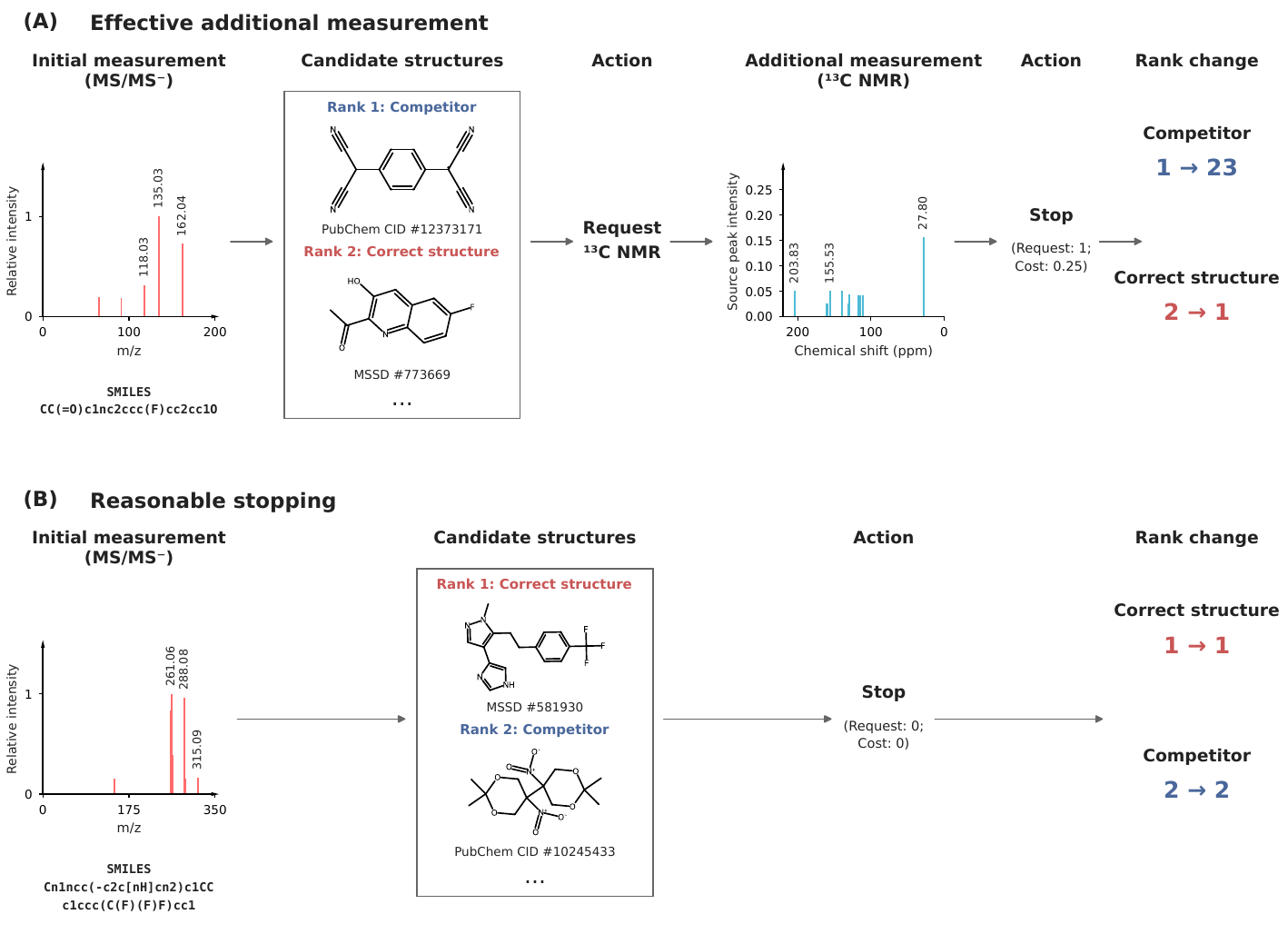}
\caption{Illustrative SI-DQN acquisition and stopping decisions from negative-ion MS/MS. \textbf{(A)} Acquisition of \textsuperscript{13}C NMR followed by stopping after the correct structure reaches rank 1. \textbf{(B)} Immediate stopping with the correct structure already ranked first. Each candidate list displays the two initially highest-ranked structures. Spectral displays and identifiers are described in Section~S1.}
\label{fig:5}
\end{figure}

To connect these population-level patterns with the spectral evidence and competing structures involved in individual decisions, two illustrative molecular cases were examined. Sample 779487 had the greatest initial structural disagreement among trajectories with a request that improved reciprocal rank. Sample 582357 had the greatest initial structural disagreement among trajectories that stopped immediately with the correct structure already ranked first. In the first case, the initial measurement was MS/MS\textsuperscript{-}. The resulting spectrum contained a base peak at m/z 135.03 and prominent signals at m/z 162.04 and 118.03, yet the correct structure was ranked second (Fig.~\ref{fig:5}A). The leading competitor differed markedly in its functional groups: the correct structure contained one carbonyl and no nitrile groups, whereas the competitor contained four nitrile groups and no carbonyl. This chemical distinction among plausible candidates provided an opportunity to assess the contribution of complementary evidence to the MS/MS-based ranking.

In response, SI-DQN requested \textsuperscript{13}C NMR, thereby adding complementary information on the carbon environments~\cite{huang2021framework,hu2024accurate}. The displayed spectrum included signals at 203.83, 155.53, and 27.80 ppm, together with the other resonances shown in Fig.~\ref{fig:5}A. After this evidence was incorporated, the correct structure moved from rank 2 to rank 1, whereas the original leading competitor fell from rank 1 to rank 23. SI-DQN then stopped acquisition after one additional measurement, incurring a normalized cost of 0.25. The substantial demotion of the competitor indicates that the added spectrum reorganized the candidate ranking beyond a simple exchange of the two leading positions. This case illustrates two aspects of the learned decision process: (1) complementary evidence was requested when the current ranking left a chemically distinct competitor unresolved; and (2) acquisition stopped immediately after that evidence was sufficient to establish the correct structure at rank 1. All four available additions yielded the same one-step net gain of +0.25, placing \textsuperscript{13}C NMR among several effective complements for this sample.

The second case illustrated the value of stopping when the initial MS/MS evidence already supported the correct ranking. The correct structure was already ranked first after the initial MS/MS\textsuperscript{-} measurement (Fig.~\ref{fig:5}B). The resulting spectrum contained prominent peaks at m/z 261.06 and 288.08 and a further signal at m/z 315.09. The correct structure and the second-ranked competitor differed strongly in composition and carbon environments: their aromatic-carbon counts were 12 and 0, and their oxygen-atom counts were 0 and 8, respectively. SI-DQN stopped immediately, with no additional measurement or acquisition cost. Each possible addition left the rank of the correct structure unchanged and incurred a cost of 0.25, corresponding to a one-step net gain of \ensuremath{-}0.25. In contrast to the first case, the candidate differences were already sufficiently resolved by the available spectral evidence. The comparison between Fig.~\ref{fig:5}A and Fig.~\ref{fig:5}B therefore shows how structural disagreement and the current spectral state jointly inform acquisition: pronounced structural differences among candidates can lead to different actions depending on what the existing evidence has established.

Together, the two cases connect the population-level improvements to the chemical distinctions and spectral evidence involved in individual decisions. In the first case, complementary evidence resolved the initial ranking ambiguity and acquisition ended after one request. In the second, the available MS/MS evidence already ranked the correct structure first, and stopping avoided acquisition with no further ranking benefit. SI-DQN combined candidate structural information with the current spectral evidence to guide both acquisition and stopping. These results extend the role of MS/MS-derived candidate rankings from an identification output to a basis for planning further characterization, with structural information helping to allocate measurement effort according to each sample's evidence needs.

\section{Conclusion}
\label{sec:conclusion}

A structure-informed deep Q-network (SI-DQN) was developed to guide complementary spectroscopic acquisition from an initial MS/MS measurement. The framework used the available spectra, candidate rankings, and chemical differences among competing structures to decide whether to obtain another spectrum or stop. On 79,462 held-out samples with simulated spectra, SI-DQN achieved Top-1 accuracies above 93\% from both MS/MS polarities with fewer than 0.35 additional measurements per sample. Additional measurement counts were reduced by 91.5--94.3\% relative to Full, and composite loss was reduced by 19.3--25.5\% relative to Rule. By adding structural information from candidate molecules to DQN, SI-DQN reduced mean additional measurement counts by 3.9\% and 8.4\% for MS/MS\textsuperscript{+} and MS/MS\textsuperscript{-}, respectively, while modestly improving Top-1 accuracy. Both benefits were observed across all three training runs. At comparable ranking uncertainty, the increase in request rate over DQN and the reduction in composite loss were greatest in groups with high structural disagreement. These findings support using candidate structural information to identify samples that can benefit from further measurement. One-step comparisons of first requests showed higher net gains for the selected measurements than for random pairing, with the requesting samples and request counts for each channel held fixed. The molecular cases illustrated how an additional spectrum resolved the initial ranking ambiguity in one sample. In the second case, immediate stopping avoided measurements that offered no further ranking benefit. This work provides a quantitative basis for using existing spectral data to decide whether further characterization is needed, which measurement to perform next, and when to stop. Future studies with experimentally measured spectra and measurement-specific acquisition costs could extend this approach to laboratory workflows.

\section*{Code and data availability}

The Multimodal Spectroscopic Dataset (MSSD v3) used in this study is publicly available at \url{https://doi.org/10.5281/zenodo.14770232}.

Reference implementations of SI-DQN and the comparison policies, executable synthetic examples, and selected aggregate results will be made publicly available on GitHub and archived on Zenodo upon acceptance.

\section*{CRediT author statement}

Bowen Gao: Conceptualization, Methodology, Software, Validation, Formal analysis, Investigation, Data curation, Visualization, Writing -- original draft, Writing -- review and editing.

Lei Zhu: Funding acquisition, Resources, Supervision, Writing -- review and editing.

Yiying Wang: Resources, Supervision, Project administration, Writing -- review and editing.

Wenjie Yu: Resources, Supervision, Project administration, Writing -- review and editing.

\section*{Generative AI disclosure statement}

During the preparation of this manuscript, the authors used ChatGPT (OpenAI) for language editing and grammar refinement to improve readability. All AI-generated content was reviewed and edited by the authors. The authors hold full responsibility for all content of the publication.

\section*{Acknowledgments}

This study was supported by the AI for Science Program, the Shanghai Municipal Commission of Economy and Informatization (2025-GZL-RGZN-BTBX-02027). The funder played no role in study design, data collection, analysis and interpretation of data, or the writing of this manuscript.

\bibliographystyle{elsarticle-num}
\bibliography{references}

\begin{thebibliography}{10}
\expandafter\ifx\csname url\endcsname\relax
  \def\url#1{\texttt{#1}}\fi
\expandafter\ifx\csname urlprefix\endcsname\relax\def\urlprefix{URL }\fi
\expandafter\ifx\csname href\endcsname\relax
  \def\href#1#2{#2} \def\path#1{#1}\fi

\bibitem{duhrkop2015searching}
K.~D{\"u}hrkop, H.~Shen, M.~Meusel, J.~Rousu, S.~B{\"o}cker, Searching
  molecular structure databases with tandem mass spectra using {CSI:FingerID},
  Proc. Natl. Acad. Sci. U.S.A. 112~(41) (2015) 12580--12585.
\newblock \href {https://doi.org/10.1073/pnas.1509788112}
  {\path{doi:10.1073/pnas.1509788112}}.

\bibitem{duhrkop2019sirius}
K.~D{\"u}hrkop, M.~Fleischauer, M.~Ludwig, A.~A. Aksenov, A.~V. Melnik,
  M.~Meusel, P.~C. Dorrestein, J.~Rousu, S.~B{\"o}cker, {SIRIUS} 4: a rapid
  tool for turning tandem mass spectra into metabolite structure information,
  Nat. Methods 16~(4) (2019) 299--302.
\newblock \href {https://doi.org/10.1038/s41592-019-0344-8}
  {\path{doi:10.1038/s41592-019-0344-8}}.

\bibitem{wang2021cfmid4}
F.~Wang, J.~Liigand, S.~Tian, D.~Arndt, R.~Greiner, D.~S. Wishart, {CFM-ID}
  4.0: More accurate {ESI-MS/MS} spectral prediction and compound
  identification, Anal. Chem. 93~(34) (2021) 11692--11700.
\newblock \href {https://doi.org/10.1021/acs.analchem.1c01465}
  {\path{doi:10.1021/acs.analchem.1c01465}}.

\bibitem{alberts2024mssd}
M.~Alberts, O.~Schilter, F.~Zipoli, N.~Hartrampf, T.~Laino, Unraveling
  molecular structure: A multimodal spectroscopic dataset for chemistry, in:
  A.~Globerson, L.~Mackey, D.~Belgrave, A.~Fan, U.~Paquet, J.~Tomczak, C.~Zhang
  (Eds.), Advances in Neural Information Processing Systems, Vol.~37, Curran
  Associates, Inc., 2024, pp. 125780--125808.
\newblock \href {https://doi.org/10.52202/079017-3996}
  {\path{doi:10.52202/079017-3996}}.

\bibitem{priessner2026mmst}
M.~Priessner, R.~J. Lewis, I.~Lemurell, M.~J. Johansson, J.~M. Goodman, J.~P.
  Janet, A.~Tomberg, Advancing structure elucidation with a flexible
  multi-spectral {AI} model, Angew. Chem. Int. Ed. 65~(2) (2026) e17611.
\newblock \href {https://doi.org/10.1002/anie.202517611}
  {\path{doi:10.1002/anie.202517611}}.

\bibitem{gao2021graphyne}
B.~Gao, M.~Sun, W.~Ding, Z.~Ding, W.~Liu, Decoration of {$\gamma$}-graphyne on
  {TiO\textsubscript{2}} nanotube arrays: Improved photoelectrochemical and
  photoelectrocatalytic properties, Appl. Catal. B: Environ. 281 (2021) 119492.
\newblock \href {https://doi.org/10.1016/j.apcatb.2020.119492}
  {\path{doi:10.1016/j.apcatb.2020.119492}}.

\bibitem{gao2021mlsynthesis}
B.~Gao, M.~Sun, Z.~Ding, W.~Liu, Machine learning-optimized synthesis of doped
  {TiO\textsubscript{2}} with improved photocatalytic performance: A multi-step
  workflow supported by designed wet-lab experiments, J. Alloys Compd. 881
  (2021) 160534.
\newblock \href {https://doi.org/10.1016/j.jallcom.2021.160534}
  {\path{doi:10.1016/j.jallcom.2021.160534}}.

\bibitem{huang2025unipoly}
Q.~Huang, Y.~Li, L.~Zhu, Q.~Zhao, W.~Yu, Unified multimodal multidomain polymer
  representation for property prediction, npj Comput. Mater. 11 (2025) 153.
\newblock \href {https://doi.org/10.1038/s41524-025-01652-z}
  {\path{doi:10.1038/s41524-025-01652-z}}.

\bibitem{gao2026robust}
B.~Gao, L.~Zhu, Y.~Wang, W.~Yu, Robust small-molecule identification from
  incomplete, degraded, and inconsistent spectra using multimodal
  mixed-condition training, arXiv:2609.14360, preprint (2026).
\newblock \href {https://doi.org/10.48550/arXiv.2609.14360}
  {\path{doi:10.48550/arXiv.2609.14360}}.

\bibitem{alberts2024infrared}
M.~Alberts, T.~Laino, A.~C. Vaucher, Leveraging infrared spectroscopy for
  automated structure elucidation, Commun. Chem. 7~(1) (2024) 268.
\newblock \href {https://doi.org/10.1038/s42004-024-01341-w}
  {\path{doi:10.1038/s42004-024-01341-w}}.

\bibitem{hu2024accurate}
F.~Hu, M.~S. Chen, G.~M. Rotskoff, M.~W. Kanan, T.~E. Markland, Accurate and
  efficient structure elucidation from routine one-dimensional {NMR} spectra
  using multitask machine learning, ACS Cent. Sci. 10~(11) (2024) 2162--2170.
\newblock \href {https://doi.org/10.1021/acscentsci.4c01132}
  {\path{doi:10.1021/acscentsci.4c01132}}.

\bibitem{beckonert2007metabolic}
O.~Beckonert, H.~C. Keun, T.~M.~D. Ebbels, J.~Bundy, E.~Holmes, J.~C. Lindon,
  J.~K. Nicholson, Metabolic profiling, metabolomic and metabonomic procedures
  for {NMR} spectroscopy of urine, plasma, serum and tissue extracts, Nat.
  Protoc. 2~(11) (2007) 2692--2703.
\newblock \href {https://doi.org/10.1038/nprot.2007.376}
  {\path{doi:10.1038/nprot.2007.376}}.

\bibitem{huang2021framework}
Z.~Huang, M.~S. Chen, C.~P. Woroch, T.~E. Markland, M.~W. Kanan, A framework
  for automated structure elucidation from routine {NMR} spectra, Chem. Sci.
  12~(46) (2021) 15329--15338.
\newblock \href {https://doi.org/10.1039/d1sc04105c}
  {\path{doi:10.1039/d1sc04105c}}.

\bibitem{mnih2015human}
V.~Mnih, K.~Kavukcuoglu, D.~Silver, A.~A. Rusu, J.~Veness, M.~G. Bellemare,
  A.~Graves, M.~Riedmiller, A.~K. Fidjeland, G.~Ostrovski, S.~Petersen,
  C.~Beattie, A.~Sadik, I.~Antonoglou, H.~King, D.~Kumaran, D.~Wierstra,
  S.~Legg, D.~Hassabis, Human-level control through deep reinforcement
  learning, Nature 518~(7540) (2015) 529--533.
\newblock \href {https://doi.org/10.1038/nature14236}
  {\path{doi:10.1038/nature14236}}.

\bibitem{hwangbo2019learning}
J.~Hwangbo, J.~Lee, A.~Dosovitskiy, D.~Bellicoso, V.~Tsounis, V.~Koltun,
  M.~Hutter, Learning agile and dynamic motor skills for legged robots, Sci.
  Robot. 4~(26) (2019) eaau5872.
\newblock \href {https://doi.org/10.1126/scirobotics.aau5872}
  {\path{doi:10.1126/scirobotics.aau5872}}.

\bibitem{watkins1992qlearning}
C.~J. C.~H. Watkins, P.~Dayan, {Q-learning}, Mach. Learn. 8~(3--4) (1992)
  279--292.
\newblock \href {https://doi.org/10.1007/bf00992698}
  {\path{doi:10.1007/bf00992698}}.

\bibitem{janisch2019classification}
J.~Janisch, T.~Pevn{\'y}, V.~Lis{\'y}, Classification with costly features
  using deep reinforcement learning, Proc. AAAI Conf. Artif. Intell. 33~(1)
  (2019) 3959--3966.
\newblock \href {https://doi.org/10.1609/aaai.v33i01.33013959}
  {\path{doi:10.1609/aaai.v33i01.33013959}}.

\bibitem{schloz2023deep}
M.~Schloz, J.~M{\"u}ller, T.~C. Pekin, W.~Van~den Broek, J.~Madsen, T.~Susi,
  C.~T. Koch, Deep reinforcement learning for data-driven adaptive scanning in
  ptychography, Sci. Rep. 13~(1) (2023) 8732.
\newblock \href {https://doi.org/10.1038/s41598-023-35740-1}
  {\path{doi:10.1038/s41598-023-35740-1}}.

\bibitem{liu2023learning}
Y.~Liu, R.~K. Vasudevan, K.~P. Kelley, H.~Funakubo, M.~Ziatdinov, S.~V.
  Kalinin, Learning the right channel in multimodal imaging: automated
  experiment in piezoresponse force microscopy, npj Comput. Mater. 9~(1) (2023)
  34.
\newblock \href {https://doi.org/10.1038/s41524-023-00985-x}
  {\path{doi:10.1038/s41524-023-00985-x}}.

\bibitem{tang2026overcoming}
X.~Tang, Y.~Zhang, X.~Deng, H.~J. Lee, D.~Zhang, Overcoming the {Nyquist} limit
  in molecular hyperspectral imaging by reinforcement learning, Adv. Intell.
  Discov. (2026) e202500240.\space\href
  {https://doi.org/10.1002/aidi.202500240} {\path{doi:10.1002/aidi.202500240}}.

\bibitem{sridharan2022deep}
B.~Sridharan, S.~Mehta, Y.~Pathak, U.~D. Priyakumar, Deep reinforcement
  learning for molecular inverse problem of nuclear magnetic resonance spectra
  to molecular structure, J. Phys. Chem. Lett. 13~(22) (2022) 4924--4933.
\newblock \href {https://doi.org/10.1021/acs.jpclett.2c00624}
  {\path{doi:10.1021/acs.jpclett.2c00624}}.

\bibitem{devata2024deepspinn}
S.~Devata, B.~Sridharan, S.~Mehta, Y.~Pathak, S.~Laghuvarapu, G.~Varma, U.~D.
  Priyakumar, {DeepSPInN} -- deep reinforcement learning for molecular
  structure prediction from infrared and {\textsuperscript{13}}{C} {NMR}
  spectra, Digit. Discov. 3~(4) (2024) 818--829.
\newblock \href {https://doi.org/10.1039/d4dd00008k}
  {\path{doi:10.1039/d4dd00008k}}.

\bibitem{efron1986bootstrap}
B.~Efron, R.~Tibshirani, Bootstrap methods for standard errors, confidence
  intervals, and other measures of statistical accuracy, Stat. Sci. 1~(1)
  (1986) 54--75.
\newblock \href {https://doi.org/10.1214/ss/1177013815}
  {\path{doi:10.1214/ss/1177013815}}.

\bibitem{field2007bootstrap}
C.~A. Field, A.~H. Welsh, Bootstrapping clustered data, J. R. Stat. Soc. B
  69~(3) (2007) 369--390.
\newblock \href {https://doi.org/10.1111/j.1467-9868.2007.00593.x}
  {\path{doi:10.1111/j.1467-9868.2007.00593.x}}.

\bibitem{vanhasselt2016deep}
H.~van Hasselt, A.~Guez, D.~Silver, Deep reinforcement learning with double
  {Q}-learning, Proc. AAAI Conf. Artif. Intell. 30~(1) (2016) 2094--2100.
\newblock \href {https://doi.org/10.1609/aaai.v30i1.10295}
  {\path{doi:10.1609/aaai.v30i1.10295}}.

\bibitem{shannon1948mathematical}
C.~E. Shannon, A mathematical theory of communication, Bell Syst. Tech. J.
  27~(3) (1948) 379--423.
\newblock \href {https://doi.org/10.1002/j.1538-7305.1948.tb01338.x}
  {\path{doi:10.1002/j.1538-7305.1948.tb01338.x}}.

\bibitem{wang2017identification}
C.~Wang, L.~He, D.-W. Li, L.~Bruschweiler-Li, A.~G. Marshall,
  R.~Br{\"u}schweiler, Accurate identification of unknown and known metabolic
  mixture components by combining {3D} {NMR} with {Fourier} transform ion
  cyclotron resonance tandem mass spectrometry, J. Proteome Res. 16~(10) (2017)
  3774--3786.
\newblock \href {https://doi.org/10.1021/acs.jproteome.7b00457}
  {\path{doi:10.1021/acs.jproteome.7b00457}}.

\end{thebibliography}

\end{document}